\documentclass[pdflatex,sn-mathphys-num,iicol]{sn-jnl}

\usepackage{graphicx}%
\usepackage{multirow}%
\usepackage{amsmath,amssymb,amsfonts}%
\usepackage{amsthm}%
\usepackage{mathrsfs}%
\usepackage[title]{appendix}%
\usepackage{xcolor}%
\usepackage{textcomp}%
\usepackage{manyfoot}%
\usepackage{booktabs}%
\usepackage{algorithm}%
\usepackage{algorithmicx}%
\usepackage{algpseudocode}%
\usepackage{listings}%
\usepackage{tkz-graph}%
\usepackage{tipa}
\usepackage{pgfplots}%
\pgfplotsset{compat=1.18}%
\usepackage{caption}%
\usepackage{subcaption}%
\usepackage{booktabs}%
\usepackage{pgfplots}%
\usepackage{pgfplotstable}%
\pgfplotsset{compat=1.18}%
\usepackage{adjustbox}
\usepackage[most]{tcolorbox}
\usepackage{xcolor}
\definecolor{mypink}{HTML}{BFBFFF}

\theoremstyle{thmstyleone}%
\theoremstyle{thmstyletwo}%

\theoremstyle{thmstylethree}%
\newcommand{\phoneme}[1]{/\textipa{#1}/}

\begin{document}

\title[Article Title]{Low resource cross-modal  alignment using HGNN to enhance speech representation}


\author*[1]{\fnm{Yannick} \sur{Yomie Nzeuhang}}\email{yynzeuhang@gmail.com}

\author[2]{\fnm{Marie} \sur{Tahon}}\email{marie.tahon@univ-lemans.fr}

\author[1,3]{\fnm{Paulin} \sur{Melatagia Yonta}}\email{paulinyonta@gmail.com}

\affil*[1]{\orgdiv{Department of computer sciences}, \orgname{University of Yaounde I}, \orgaddress{\street{Street}, \city{Yaounde}, \postcode{812}, \country{Cameroon}}}

\affil[2]{\orgdiv{IRD}, \orgname{UMMISCO}, \orgaddress{\street{Street}, \city{Bondy}, \postcode{F-93143}, \country{France}}}

\affil[3]{\orgdiv{LIUM}, \orgname{ Le Mans Université}, \orgaddress{\street{Avenue Olivier Messiaen}, \city{Le Mans}, \postcode{72085}, \country{France}}}


\abstract{

Speech-text space alignment is a multimodal representation learning method consisting to map different speech and text into a shared representation space, leading to enrichment of the representation of each modality. Proposed architectures, such as SAMU-XLSR, typically follow a student/teacher framework, with the goal of fine-tuning an audio encoder to produce representations that closely match those of the text. In this way a speech representation is semantically enriched. However, such systems generally require large amounts of training data and considerable computational resource, making them difficult to apply to low resources languages under frugal constraints. The present work proposes a data-efficient space alignment method based on Heterogeneous Graph Neural Networks and link prediction. The core idea is to leverage message passing to explicitly transfer information from the text modality to the speech modality, thereby reducing the need for large training datasets and intrinsically enriching the acoustic representations, all in a more interpretable manner. Although thoroughly explored for high-resource languages, word-level tasks in speech remain relevant for certain low-resource languages. Therefore, we conducted experiments on speech-text alignment at the word level using the TIMIT (English) dataset and Yemba (a Cameroonian language). Our approach yields results comparable to those of SAMU-XLSR, a state-of-the-art method, and even surpasses it for the Yemba language in the task of word retrieval, while using far fewer resources, demonstrating its power, frugality, and efficiency.}




\maketitle

\section{Introduction}
\label{sec:in}

Representation learning~\cite{emb} is a process in machine learning whereby algorithms extract meaningful patterns from raw data, generating representations that are more suitable for downstream tasks. 
Recently, multimodal representation learning has emerged as a strategy to strengthen the quality of representations, as it mirrors the way humans process interdependent modalities such as vision and language, or speech and text. Among the various approaches in this field, cross-modal alignment aims to map different modalities into a shared representation space, such that representations of different modalities corresponding to the same semantic object lie close to each other, which, as a result, enriches the representations of each modality. This strategy has shown great promise in image processing, yielding significant improvements in downstream tasks such as image captioning~\cite{clip,EAMA}. Now it is gaining traction in speech processing and for speech-text alignment. In particular, recent studies~\cite{speech_mining,SAMU-XLSR, SONAR} propose to encode speech signals into fixed-sized representations that are optimized to minimize cosine distance with pre-trained frozen text embeddings which lead to semantically enriching the speech representation.

Such enriched representations are expected to be particularly useful in low-resource settings, where the quality of representations is crucial \cite{wav2vec, xlsr-53}, given the limited amount of data available for training downstream tasks. However, although effective, current alignment approaches typically require fine-tuning large acoustic encoders with billions of parameters on vast amounts of parallel data. In the context of low-resource languages, such large quantities of data are not available. Furthermore, the computational resources required to train these models are often lacking. Therefore, there is a need for lightweight approaches tailored to low-resource and frugal settings, which is the purpose of this paper. Indeed in this paper, we propose a frugal cross-modal alignment method to obtain enriched speech representations in low resource constraint. This approach consists in learning a cross-modal representation where the alignment is modelled as a link prediction problem in a heterogeneous graph. In this graph, nodes represent elements from distinct modalities (\textit{e.g.}, acoustic and linguistic). 
Our approach leverages a heterogeneous graph neural network (GNN), where information is exchanged between acoustic and text-based nodes via message passing mechanisms.
GNNs have recently demonstrated success across numerous audio domains, including emotion recognition~\cite{graphMFT, Shirian, liu_emotion}, speaker diarization~\cite{dia1,dia_2}, and audio/speech classification~\cite{Castro_audio, Shirian, Shilei_few_shot}, highlighting their versatility and effectiveness. 
To the best of our knowledge, this is the first cross-modal representation learning framework based on GNN applied to speech-text space alignment.

Although thoroughly explored for high-resource languages, word-level tasks \cite{Menon2018FeatureEF,Jacobs2021MultilingualTO,Jacobs2021AcousticWE,Kamper2020ImprovedAW,vanderwesthuizen23_sigul} in speech remain relevant for certain low-resource languages. Therefore our experiments focus specifically on word-level alignment, measuring the enrichment of the acoustic word representation by standard clustering metrics on two languages: English as a reference language and Yemba language spoken in Cameroon as a low resource language base. Although our work focuses on the enrichment of speech representations, we also assess the effectiveness of our approach for cross-modal word alignment through a word retrieval task.
Through this work, we make the following methodological contributions:
\begin{itemize}
    \item We propose a novel and lightweight framework for cross-modal alignment in a shared representation space, based on heterogeneous GNN.
    \item We demonstrate its effectiveness in aligning speech and text representations at the word level.
    \item We introduce a phonological alignment task. This task leverages a structured phonological representation of words, leading to more interpretable acoustic embeddings after alignment. The experiment also demonstrate the power of this alignment in enriching the acoustic representation.
\end{itemize}
The code is available on GitHub for reproducibility concerns\footnote{\url{https://github.com/terrencetao/Acoustic_Linguistic_GNN/tree/exp_link_pred_binary_articulatoire}}

The rest of this paper is organized as follows: Section~\ref{sec:rel} discusses related work. Section~\ref{sec:method} presents the proposed graph-based model and details how acoustic and linguistic cues are integrated. Section~\ref{sec:exp} outlines our experimental   and Section \ref{RD} presents the results. Finally, Section~\ref{sec:con} concludes the paper and outlines future directions.

\section{Related works}
\label{sec:rel}

\subsection{Graph neural networks}
A growing field of research explores the use of Graph Neural Networks (GNNs) in image~\cite{GNN_img,Gnn_img_2}, text~\cite{gnn_text} or audio processing~\cite{Shirian,SHARC,dialogueGCN}. Their popularity is steadily increasing, driven in part by their explainability potential and the flexibility induced by the message passing principle. GNNs \cite{gnn_1} are a class of neural networks designed specifically for handling data organized in graph structures. A graph is defined as a tuple consisting of a set of nodes and a set of edges between them. Without modifying the structure of the graph, GNNs capture the dependencies within the graph by updating node embeddings through a message-passing mechanism between its nodes~\cite{MPNN}. Indeed, to update the information contained in a node $u$, GNNs aggregate information from the neighbouring nodes and combine it into a final representation. 
Formally let $h^{k}_{u}$ represent the embedding of node $u$ at the $k-th$ iteration of the message-passing process, and $N(u)$ a set of neighbours of $u$, $h^{k+1}_{v}$ is defined in eq.~\ref{eq:message-passing}.
\begin{equation}\label{eq:message-passing}
   h_u^{(k+1)} = \text{UPD}^{(k)} \left( h_u^{(k)}, \text{AGG}^{(k)} \left( \{ h_v^{(k)} \mid v \in N(u) \} \right) \right)
\end{equation}

\noindent Where  \( \text{AGG}^{(k)} \) combines the embeddings of \( N(u) \) and \( \text{UPD}^{(k)} \) updates the embeddings of \( u \) using its current embedding and aggregated neighbour information. 

\subsection{Multimodal representation learning}
 In the context of text-audio space alignment, a common approach is to minimize a distance-based loss between audio and text embeddings~\cite{speech_mining} under a teacher-student paradigm.
More precisely in recent literature, the goal is to increase the cosine similarity between the output of a fixed text teacher encoder (LASER~\cite{laser}) and that of a trainable speech student encoder (XLSR~\cite{xlsr-53}). 

Building on this idea, SAMU-XLSR~\cite{SAMU-XLSR} extended previous approach by replacing LASER with LaBSE~\cite{LaBSE}, a multilingual text encoder, as the teacher. They demonstrated the effectiveness of this method across 25 languages such French, English, German and downstream tasks such retrieval and sequence-to-sequence generation. SONAR~\cite{SONAR} is one of the most recent advancement in this area, which differs from earlier works by minimizing the mean squared error (MSE) loss between sentence embeddings of transcriptions and those of speech. Moreover, they introduce their own text encoder as the teacher.
While these approaches have proven effective, they focus exclusively on semantic alignment and do not explicitly model knowledge transfer between the text and audio modalities. 

In this work, we propose a more interpretable and general framework for enriching acoustic embeddings using text-based information via GNNs. Our approach follows the teacher-student paradigm, where the text encoder serves as a fixed teacher. However, unlike prior work, we leverage the message-passing mechanisms of GNNs to facilitate explicit knowledge transfer between modalities. 

In addition to semantic alignment, we also introduce a novel phonetic alignment task. This task focuses on phonological similarity rather than purely semantic content, and is particularly relevant for linguistically grounded applications such as speech synthesis and language documentation.

\subsection{Multimodal representation learning with GNN in audio}
In the context of GNNs, the goal of multimodal representation is to build heterogeneous graphs that capture both the information within each modality and the interactions between modalities. An example is MMGCN \cite{MMGCN}, a method that leverages audio, text, and image modalities for emotion recognition using GNN. 
For each conversation, a graph is constructed in which the nodes represent different modalities. Nodes belonging to the same modality within a conversation are connected (capturing intra-modal information), while each node is also linked to nodes from other modalities corresponding to the same utterance (capturing cross-modal interactions). However, MMGCN treat all modalities at the same level, and does not sufficiently address the heterogeneity of multimodal data, \textit{i.e.} the interaction between these modalities. To address this, GraphMFT \cite{graphMFT} proposes a more nuanced multimodal fusion technique for emotion recognition. Instead of combining all three modalities at once, GraphMFT constructs three separate graphs, each containing two modalities. In these graphs, two types of edges are used: intra-modal edges, which connect nodes within the same modality based on temporal context (\textit{i.e.}, linking past or future contextual nodes) and inter-modal edges, which connect nodes from different modalities but corresponding to the same utterance.
These graphs are processed through specialized Graph Attention Networks (GATs)\cite{gat}, and the resulting representations are concatenated pairwise, resulting in three vectors. These vectors are further combined through linear projections to form the final representation. This design allows the model to capture intrinsic information within each modality (via intra-modal edges) and interactions between modalities (via inter-modal edges). 

Despite recent advances, the literature on GNNs, especially in multimodal approaches, remains limited.
To the best of our knowledge, no articles have been published on the use of GNNs to model acoustic and linguistic information in a multimodal framework.
To bridge this gap, the present work introduces a Heterogeneous Graph Neural Network (HGNN) to learn a shared text-acoustic representation space at the word level. The proposed method models this space alignment task as a link prediction task in a heterogeneous graph.


\section{Proposed Method}
\label{sec:method}
The main idea behind our work is to enrich the acoustic representation with linguistic information. To achieve this objective, our method must allow the transfer of information from one modality to another in a teacher-student paradigm. 
 
\begin{figure*}[t]
    \centering
    \begin{subfigure}{0.46\textwidth}
        \centering
        \includegraphics[width=\linewidth]{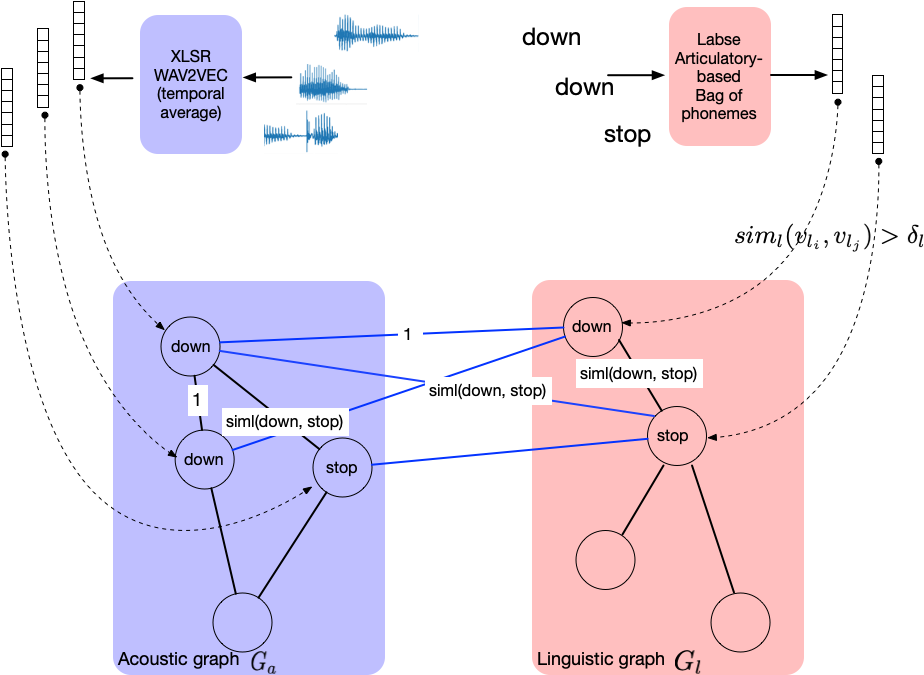}
         \caption{Construction of the heterogeneous graph. Blue denotes acoustic nodes, red denotes transcription nodes}
        \label{fig:gcn_architecture}
    \end{subfigure}
    \hfill
    \begin{subfigure}{0.46\textwidth}
        \centering
        \includegraphics[width=0.8\linewidth]{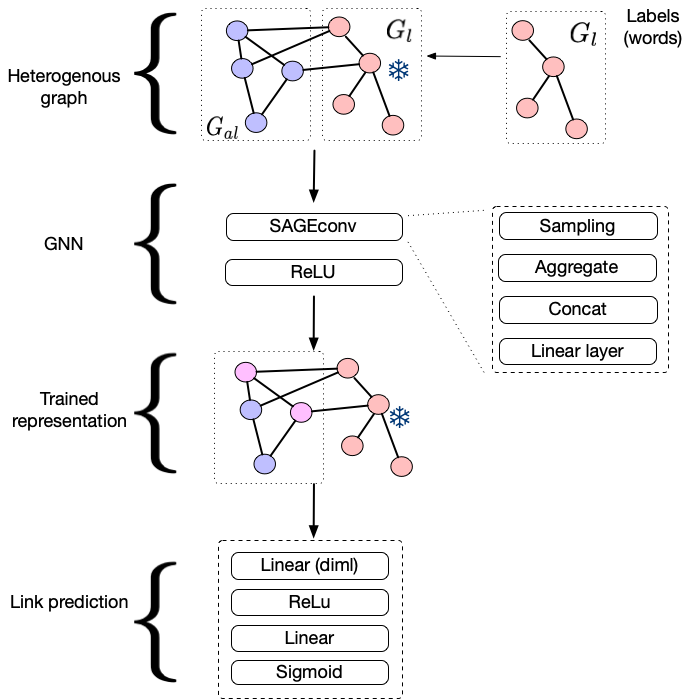}
       
        \caption{Training Architecture of the graph Convolutional neural network}
        \label{fig:gcn}
    \end{subfigure}

    \caption{Overview of the heterogeneous graph construction protocol and the GCN training architecture.}
    \label{fig:enter-label}
\end{figure*}

We propose to construct a heterogeneous graph, illustrated in Figure~\ref{fig:gcn_architecture}, whose nodes are made up of spoken words and their transcriptions, and then to learn the final representation of words with the GNN paradigm. By virtue of its message-passing learning principle, GNN allows the transfer of information between the linguistic and acoustic nodes of the graph.



\subsection{The heterogeneous graph}
Before training a GNN on this data, we need to find a principle for constructing a graph from this data. Our heterogeneous graph is composed of three sub-graphs named (i) $G_a$ acoustic sub-graph, where nodes are representations of spoken words, (ii) $G_l$ linguistic sub-graph, where nodes are representations of word transcriptions, and (iii) $G_{al}$ acoustic-linguistic sub-graph where the nodes are the union of the two previous sub-graphs.

\subsubsection{The acoustic sub-graph}

Given a collection of \( N \) audio samples for training, we construct an undirected weighted graph \( G_{a} = (V_{a}, E_{a}) \) to model relationships among acoustic instances, where \( V_{a} \) is the set of nodes (audio samples), and \( E_{a} \) is the set of edges connecting them. Let \( v_{a_i}, v_{a_j} \in V_{a} \) be two nodes, and let \( SIM_{a}(v_{a_i}, v_{a_j}) \) denote a similarity measure between their acoustic representations. The acoustic representation can be any valid feature type. For our experiments, we use features extracted from pre-trained speech representations namely XLSR~\cite{xlsr-53} and Wav2Vec2~\cite{wav2vec} at the word level.

Let \( T(v_{a_j}) \) denotes the set of nodes \( v_{a_j} \in V_{a} \) that share the same transcription as \( v_{a_i} \). For every node \( v_{a_j} \in T(v_{a_i}) \), an edge \( \{v_{a_i}, v_{a_j}\} \) is added with a weight of 1 to represent a strong intra-class connection.

For each node \( v_{a_k} \in V_{a} \setminus T(v_{a_i}) \) (\textit{i.e.}, having a different transcription from \( v_{a_i} \)), an edge \( \{v_{a_i}, v_{a_k}\} \) is added, weighted by $SIM_l(v_{a_i}, v_{a_k})$ the similarity between the transcriptions of \( v_{a_i} \) and \( v_{a_k} \) in the linguistic sub-graph. This ensures that acoustically different words are connected in a way that reflects their linguistic proximity. Since we want smooth communication of information without redundancy, once the graph has been constructed, we use the maximum spanning tree~\cite{Yassin2023} to filter the graph and obtain only the main structure.

 \subsubsection{The linguistic sub-graph}
 \label{sub-sec:lg}
Given a collection of \( M \) transcribed words, we construct an undirected weighted graph \( G_{l} = (V_{l}, E_{l}) \).
$v_{l_{i}}$ and $v_{l_{j}} \in V_{l}$ are two nodes represented by their linguistic embeddings, $SIM_{l}(v_{l_{i}},v_{l_{j}})$ is defined as a similarity measure between these nodes. $\delta_{l} \in \mathbb{R_{+}}$, a hyperparameter used as a threshold to restrict the number of edges in $G_{l}$.$\delta_{l} \in \mathbb{R_{+}}$, our filter allows us to retain only those edges that are linguistically relevant, but also relevant in terms of weight, which will be used in the other two graphs, as shown in Figure \ref{fig:enter-label}.

For each  node $v_{l_{i}}  \in V_{l}$ an edge \(\{v_{l_{i}},v_{l_{j}} \}\) is added  if $SIM_{l}(v_{l_{i}},v_{l_{j}})> \delta_{l}$, the weight of \(\{v_{l_{i}},v_{l_{j}} \}\) is set to $SIM_{l}(v_{l_{i}},v_{l_{j}})$.
To obtain word embeddings, we use standard text encoders such as LaBSE~\cite{LaBSE}. 
This choice is motivated by the fact that LaBSE is supposed to be multilingual and provides good representations for many languages. However, multilingual representations are known not to be robust for some languages~\cite{Chimoto2022VeryLR}.
Therefore, in addition to these pretrained models, we propose two phonetic representations specifically designed for phonetic speech-text alignment tasks:

\begin{enumerate}[(i)]
    \item \textbf{Articulatory-based representation:} A word is represented as a sequence of phonemes, where each phoneme is encoded as a vector of articulatory features (\textit{e.g.} voiced, rounded, etc.) with 24 dimensions~\cite{panphon}. Each dimension indicates the presence (1), the absence (-1), or the non-relevance (0) of a given feature. The final word embedding is then obtained by concatenating the articulatory feature vectors of its constituent phonemes.
    
    \item \textbf{Bag-of-phonemes representation:} This representation is inspired by the traditional ``bag-of-words'' model but operates at the phoneme level. We first build a vocabulary by transcribing a set of \( M \) words into their phonetic forms and collecting the set of unique phonemes. Each word is then represented as a vector of phoneme occurrences within that word, using this vocabulary. This results in a sparse vector, where each dimension corresponds to the count of a specific phoneme. 
    
    As an example, we consider a subset of words from the Google Speech Commands dataset~\cite{speechcommandsv2} and their phonetic transcriptions: \textbf{down}: \phoneme{daU"n}, \textbf{go}: \phoneme{goU}, \textbf{left}: \phoneme{lEft}, \textbf{no}: \phoneme{noU}, \textbf{right}: \phoneme{raIt}, \textbf{stop}: \phoneme{st6p}, \textbf{up}: \phoneme{Vp}, \textbf{yes}: \phoneme{jEs}. 
    Table~\ref{tab:bagofph} presents the  bag-of-phoneme vectors generated for each word using this method.
\end{enumerate}

 \begin{table*}
 \small
\centering
\caption{Bag-of-phonemes representations of each word from the ``Google Speech Commands''.}

    \begin{tabular}{r|*{17}{c}}

    \toprule
    \text{} & a & d & f & j & l & n & o & p & r & s & t & \text{\textipa{6}} & \text{\textipa{E}} & \text{\textipa{g}} & \text{\textipa{I}} & \text{\textipa{U}} & \text{\textipa{V}} \\
    \midrule
    \text{down} & 1 & 1 & 0 & 0 & 0 & 1 & 0 & 0 & 0 & 0 & 0 & 0 & 0 & 0 & 0 & 1 & 0 \\

    \text{go} & 0 & 0 & 0 & 0 & 0 & 0 & 1 & 0 & 0 & 0 & 0 & 0 & 0 & 1 & 0 & 1 & 0 \\

    \text{left} & 0 & 0 & 1 & 0 & 1 & 0 & 0 & 0 & 0 & 0 & 1 & 0 & 1 & 0 & 0 & 0 & 0 \\

    \text{no} & 0 & 0 & 0 & 0 & 0 & 1 & 1 & 0 & 0 & 0 & 0 & 0 & 0 & 0 & 0 & 1 & 0 \\

    \text{right} & 1 & 0 & 0 & 0 & 0 & 0 & 0 & 0 & 1 & 0 & 1 & 0 & 0 & 0 & 1 & 0 & 0 \\

    \text{stop} & 0 & 0 & 0 & 0 & 0 & 0 & 0 & 1 & 0 & 1 & 1 & 1 & 0 & 0 & 0 & 0 & 0 \\

    \text{up} & 0 & 0 & 0 & 0 & 0 & 0 & 0 & 1 & 0 & 0 & 0 & 0 & 0 & 0 & 0 & 0 & 1 \\

    \text{yes} & 0 & 0 & 0 & 1 & 0 & 0 & 0 & 0 & 0 & 1 & 0 & 0 & 1 & 0 & 0 & 0 & 0 \\
    \bottomrule
    
\end{tabular}
\label{tab:bagofph}
\end{table*}

\subsubsection{ The acoustic-linguistic sub-graph}
 \label{subsec:hetero}
Given a collection of \( N \) audio speech samples and their corresponding \( M \) transcribed words, we define the undirected acoustic-linguistic sub-graph \( G_{al} = (V_{al}, E_{al}) \), which connects nodes from the previous acoustic \( G_a \) and linguistic \( G_l \) graphs. Here, \( V_{al} = V_{a} \cup V_{l} \) represents the union of acoustic and linguistic nodes, and \( E_{al} \) denotes the set of edges connecting acoustic nodes \( v_{a_i} \in V_a \) with linguistic nodes \( v_{l_j} \in V_l \).

To construct the set \( E_{al} \), we add an edge \( \{v_{a_i}, v_{l_j}\} \in E_{al} \) with a weight of 1 if \( v_{l_j} \) is the transcription corresponding to speech sample represented by \( v_{a_i} \). Otherwise, the edge \( \{v_{a_i}, v_{l_j}\} \) is assigned a weight equal to the similarity between the transcriptions of \( v_{a_i} \) and \( v_{l_j} \), as computed within the linguistic sub-graph \( G_l \). According to the loss function \ref{eq:loss}, each link in this graph will be used during training. The more we have, the better the model will be able to learn.

 \subsection{Learning acoustic representation}
 \subsubsection{Graph convolutional neural model }
  \label{subsubsec:model}

Most GNNs are transductive, \textit{i.e.} all nodes in the graph must be present during training, and once the model is trained, its structure is fixed. Consequently, with transductive approaches, it is not possible to insert a new node in the graph during inference.
To cope with this issue, GraphSAGE \cite{GraphSAGE} introduces a general framework for GCN with an inductive property.
The inductive approach learns message passing functions that generate embeddings by sampling and aggregating features from the local neighborhood of a node. 
For a GraphSAGE model, the convolutional layers (SAGEconv) have 4 message-passing algorithm options, which differ mainly in the \textit{aggregator} defined in eq.~\ref{eq:message-passing}. In our experimentation, we use the simplest form called \textit{average aggregator} defined in eq.~\ref{eq:graphsage} where $h_u^k$ is the embedding of node $u$ at iteration $k$, $\sigma$ is a non-linear function (ReLU or sigmoid), $W_k$ is a learned matrix, and $N(u)$ the neighbors of node $u$.

We use a GNN of type GraphSAGE which consists of one convolutional block (SAGEconv\footnote{\url{https://pytorch-geometric.readthedocs.io/en/latest/generated/torch_geometric.nn.conv.SAGEConv.html}} described in Figure~\ref{fig:gcn}) and a ReLU activation.

\begin{equation}\label{eq:graphsage}
h_u^{k+1} = \sigma \left( W_{k} \cdot \text{MEAN} \left( \left\{ h_u^{k} \right\} \cup \left\{ h_v^{k} \mid \forall v \in N(u) \right\} \right) \right)
\end{equation}


  
  
\subsubsection{Overall Objective Function}

Our training objective combines three complementary loss terms (see eq.~\ref{eq:loss}) designed to (i) predict the strength of cross-modal associations, (ii) align acoustic and word embeddings, and (iii) preserve local acoustic similarity. 
Hyperparameters $\alpha$ and $\beta$ control the balance between the alignment and regularization terms in the total loss.

The overall loss is defined as:


\begin{equation}\label{eq:loss}
\mathcal{L}_{\text{total}}
= \mathcal{L}_{\text{reg}}
+ \alpha \mathcal{L}_{\text{contrast}}
+ \beta \mathcal{L}_{\text{acoustic}}.
\end{equation}

In the followings, $\mathcal{E}$ is the set of all word-acoustic node pairs $(w_i, a_j)$, while $\mathcal{P}$ is the set of pairs for which $w_i$ is the transcription of the speech sample $a_j$.


\paragraph{Binary Regression Loss}

The first term, $\mathcal{L}_{\text{reg}}$, defined by eq.~\ref{eq:lreg}, is a binary cross-entropy (BCE) loss that learns to predict the existence and strength of links between word and acoustic nodes. Let:

\begin{itemize}
    \item \( y_{i,j} \in [0,1] \) denotes the observed link weight between word node \( w_i \) and acoustic node \( a_j \).
    \item 
$p_{i,j} = \sigma\left(\text{MLP}\left[\mathbf{{h}_{w_i}} \,\Vert\, \mathbf{{h}_{a_j}} \,\Vert\, \mathbf{{h}_{w_i}} - \mathbf{{h}_{a_j}} \,\Vert\, \mathbf{{h}_{w_i}} \odot \mathbf{{h}_{a_j}}\right]\right)$

is the predicted link prediction probability computed by applying a multi-layer perceptron (MLP) followed by a sigmoid activation to the concatenated features. The MLP consists of two linear layers and a ReLU activation as illustrated in Figure~\ref{fig:gcn}.
\end{itemize}
{\small
\begin{equation}\label{eq:lreg}
    \mathcal{L}_{\text{reg}} = -\frac{1}{|\mathcal{E}|} \sum_{(i,j) \in \mathcal{E}} \left[ y_{i,j} \log(p_{i,j}) + (1 - y_{i,j}) \log(1 - p_{i,j}) \right]
\end{equation}
}

\paragraph{Heterogeneous Contrastive Loss (Word-Acoustic)}

To align the acoustic and word representations in a shared representation space, we use an InfoNCE contrastive loss~\cite{InfoNCE}. Let:

\begin{itemize}
    \item $h_{w_i}, h_{a_j}$ be $\ell_2$-normalized embeddings,
    \item $\tau=0.07$ be a temperature hyperparameter set empirically.
\end{itemize}

The loss encourages each positive pair to have a higher similarity than all negatives in the batch:

{\small
\begin{equation}
    \mathcal{L}_{\text{contrast}} = -\frac{1}{|\mathcal{E}|} \sum_{(i,j)\in \mathcal{P}} \log \left( \frac{\exp(h_{w_i}^T h_{a_j} / \tau)}{\sum_{(i,j)\in \mathcal{E}} \exp(h_{w_i}^T h_{a_j} / \tau)} \right)
\end{equation}
}
\paragraph{\textbf{Homogeneous Contrastive Loss (Acoustic-Acoustic)}}

To preserve local similarity between acoustically similar nodes (\textit{e.g.}, different utterances of the same word), we add a homogeneous contrastive term. Let:

\begin{itemize}
    \item $\mathcal{P}_a = \{(a_k, a_k^+)\}_{k=1}^K$ be positive acoustic-acoustic pairs (similar words),
    \item $h_{a_k}, h_{a_k^+}$ be normalized acoustic embeddings.
    \item $\tau$ is a temperature parameters equal to the previous one ($\tau = 0.07$)
\end{itemize}

This loss is also based on the InfoNCE formulation:
\begin{equation}
    \mathcal{L}_{\text{acoustic}} = -\frac{1}{K} \sum_{k=1}^K \log \left( \frac{\exp(h_{a_k}^T h_{a_k^+} / \tau)}{\sum_{m=1}^K \exp(h_{a_k}^T h_{a_m} / \tau)} \right)
\end{equation}

\begin{itemize}
    \item \( y_{i,j} \in [0,1] \) denotes the observed weight of the link between the lexical node \( w_i \) and the acoustic node \( a_j \)
    \item 
    {\small
    $
    p_{i,j} = \sigma\left(
    \text{MLP}\left[
    \mathbf{h}_{w_i} \,\Vert\, 
    \mathbf{h}_{a_j} \,\Vert\, 
    (\mathbf{h}_{w_i} - \mathbf{h}_{a_j}) \,\Vert\, 
    (\mathbf{h}_{w_i} \odot \mathbf{h}_{a_j})
    \right]
    \right)
    $}
    represents the predicted probability of link existence, obtained by applying a multilayer perceptron (MLP), followed by a sigmoid activation, to the concatenation of the latent representations. The MLP consists of two linear layers separated by a ReLU activation, as illustrated in Figure~\ref{fig:gcn}.
\end{itemize}

\section{Experimental protocol}
\label{sec:exp}
The main objective of our experiments is to demonstrate the effectiveness of our alignment approach in enriching word-level acoustic representations, achieving performance comparable to state-of-the-art methods, notably SAMU-XLSR\cite{SAMU-XLSR}, while requiring fewer resources. The acoustic word representations produced are evaluated along two intrinsic axes: an intrinsic unimodal and a cross-modal axis. On the intrinsic unimodal level, we measure the intra-class and inter-class similarity of word-level acoustic representations, which respectively assess the compactness of the representations of a given word across its different occurrences, and their separability with respect to the representations of other words. A high intra-class similarity combined with a low inter-class similarity reflects an acoustic space well structured. On the intrinsic cross-modal level, we evaluate the representations through a word retrieval task (a word-level adaption of the retrieval task used in SAMU-XLSR\cite{SAMU-XLSR} paper), which consists, given a text word, in retrieving the acoustic words that correspond to it semantically. The comparison is performed between the words retrieved on the acoustic side and those retrieved on the text side for the same query, which allows us to directly assess the quality of the cross-modal alignment produced by each method, rather than solely the internal organization of the acoustic space. 

Alongside this main objective, we also assess the impact of different initial representations on the performance of our method, including LaBSE, bag-of-phonemes, and articulatory-based representations for text, as well as wav2vec and XLSR for speech.
 
\subsection{Dataset Presentation}
We evaluate our method on two datasets: TIMIT and a dataset in Yemba. In addition to a low resource language, we choose to include English data to ensure the reproducibility of the method.
For both datasets, we randomly selected 80\% of the data to train the HGNN and the rest to evaluate the link prediction with regression metrics. Retrieval and enrichment evaluation is realized with additional lists of words under two settings: transductive; where test words are those used to train the graph, and inductive where test words are not seen during training.
\begin{itemize}
    \item \textbf{TIMIT} \cite{timit} dataset contains a large set of sentences pronounced by 630 speakers. To simulate a low resource scenario, we selected words whose occurrences varies between 30 and 100 ($N_w$). This selection consists of $W=109$ unique words and 5,389 utterances. 
In the inductive setting, two disjoint sets are manually created: one with semantically similar words and another with phonetically similar words, enabling a finer evaluation of alignment.
\item \textbf{Yemba dataset (YD)} \cite{KanaAzeuko2024} contains 8,031 recordings of $W=60$ unique words pronounced by 69 native speakers. For the inductive setting, 100 additional words are randomly extracted from YembaTones~\cite{kenfack2023yembatones}. Unlike TIMIT, no semantic/phonetic split is available due to the lack of linguistic expertise.
\end{itemize}


All audio samples have been resampled at 16kHz. Then acoustic embeddings are extracted using two pretrained models, then averaged with time. The pre-trained models are XLSR~\cite{xlsr}\footnote{facebook/wav2vec2-xls-r-300m} (1024 dimensions) and Wav2Vec 2.0~\cite{wav2vec}\footnote{facebook/wav2vec2-base-960h} (768 dimensions); wav2vec being trained on english, we only used it on the TIMIT, not on the Yemba..  
For the linguistic modality, we compare three types of representations: LaBSE~\cite{LaBSE} embeddings, bag-of-phonemes (see sec.~\ref{sub-sec:lg}), and articulatory-based vectors (see sec.~\ref{sub-sec:lg}). To extract the articulatory features we used PanPhon~\cite{panphon} library of python 3.10.

\subsection{Evaluations metrics}

    \label{subsec:expp}

The experiments are conducted on both TIMIT and YD datasets, in the two evaluation contexts described in section~\ref{subsubsec:model}: a transductive context where words are visible during training, and an inductive context involving words outside the vocabulary.
Our evaluation protocol includes two tasks: (i) representation enrichment, and (ii) cross-modal alignment.

\subsubsection{Representation enrichment} 
Enrichment is measured via intra- and inter-similarity between acoustic vectors using cosine similarity before and after the cross-modal training. These measures are inspired by the intrinsic evaluation process for word embeddings \cite{ghannay-etal-2016-word}, which aims to evaluate the consistency between the semantic link between two words and the similarity of the associated embeddings. In our case, we sought to determine the degree of proximity between the audio representations of the same word (intra-similarity) and the representations of different words (inter-similarity). Specifically, we compute: 
\begin{itemize}
    \item \textbf{Intra-similarity} The average cosine similarity between acoustic nodes sharing the same target word:  
{\footnotesize
\begin{equation}\label{eq:intra}
S_{\text{intra}} =
\frac{1}{|W|} \sum_{w \in W}
\frac{1}{|N_w|(|N_w|-1)}
\sum_{\substack{i,j \in N_w \\ i \neq j}}
simcos(v_i, v_j),
\end{equation}
}
where $W$ is the set of unique words, $N_w$ is the set of nodes (acoustic realizations) for word $w$, and $v_i$ represents the embedding vector of node $i$.  

\item \textbf{Inter-similarity} The average cosine similarity between nodes with different target words:  
{\footnotesize
\begin{equation}\label{eq:inter}
S_{\text{inter}} =
\frac{1}{|W|(|W|-1)}
\sum_{\substack{w_1 \neq w_2}}
\frac{1}{|N_{w_1}||N_{w_2}|}
\sum_{\substack{i \in N_{w_1}\\ j \in N_{w_2}}}
simcos(v_i, v_j)
\end{equation}
}
\end{itemize}

Good representations should exhibit high intra-similarity (acoustic realizations of the same word should be close in the embedding space) and low inter-similarity (different words should be well-separated).

\subsubsection{Cross-modal alignment} Alignment evaluates the degree to which retrieval in the acoustic space agrees with retrieval in the text space. We based our metric on the classic retrieval one \cite{SAMU-XLSR}, whose principle is to measure the model's ability to retrieve relevant information for a given query. However, unlike the latter, where they already have the ground truth, we construct it for each word query.
Given a query word $w$, we consider two retrieval results:

\begin{itemize}
    \item \textbf{Text-based retrieval (ground truth):} is the set of closest neighbors of word $w$ in the linguistic space. It serves as the ground-truth ranking in the text modality.
    \begin{equation}
        R^{\text{text}}_{k}(w) = \operatorname{TopK}_{u \in \mathcal{T}} simcos(t_w, t_u),
    \end{equation}
    where $t_w$ denotes the linguistic embedding of word $w$, $\mathcal{T}$ is the set of all text embeddings, and $\operatorname{TopK}$ returns the $k$ most similar items based on cosine similarity.

    \item \textbf{Acoustic-based retrieval:} is the set of closest neighbours of word $w$ in the cross-modal space. It corresponds to retrieving words from the acoustic modality using the same text query.
    \begin{equation}
        R^{\text{acoustic}}_{k}(w) = \{\ell(v) \mid v \in \operatorname{TopK}_{v \in V} simcos(t_w, v)\},
    \end{equation}
    where $V$ is the set of acoustic embeddings, and $\ell(v)$ maps an acoustic vector to its corresponding word label. 
    \end{itemize}

The quality of cross-modal alignment is assessed by comparing $R^{\text{text}}_{k}(w)$ and $R^{\text{acoustic}}_{k}(w)$ using two complementary metrics:

\begin{itemize}
    \item \textbf{Word Overlap at $k$ (WO@$k$)}.

Measures the proportion of exact word matches between the two retrieved sets:
\begin{equation}
    \text{WO@k}(w) = \frac{|R^{\text{text}}_{k}(w) \cap R^{\text{acoustic}}_{k}(w)|}{k}.
\end{equation}

\item \textbf{Phonetic Overlap at $k$ (PO@$k$)}.
To account for phonetic similarity, we define
\begin{equation}
    \text{PSim}(u,v) = 1 - \frac{\text{Levenshtein}(u_{p},v_{p})}{\max(|u_{p}|,|v_{p}|)},
\end{equation}
where $u_{p}$ and $v_{p}$ are phonetic spelling of words $u$ and $v$ respectively, and similarity is considered valid when the phonetic similarity is greater than $\delta_{p}$ which we set to 0.5.

For each acoustic-retrieved word $v \in R^{\text{acoustic}}_{k}(w)$, we check whether there exists at least one text-retrieved word $u \in R^{\text{text}}_{k}(w)$ that is phonetically similar. Each acoustic word contributes at most once, even if multiple text words are similar. Formally:

{\footnotesize
\begin{equation}
    \text{PO@k}(w) = \frac{1}{k} \sum_{v \in R^{\text{acoustic}}_{k}(w)} 
    \mathbf{1}\left(\max_{u \in R^{\text{text}}_{k}(w)} \text{PSim}(u,v) \geq \delta_{p} \right).
\end{equation}
}

\end{itemize}

For both WO@$k$ and PO@$k$, we report averages across all query words $Q$:

\begin{equation}
\begin{aligned}
\overline{\text{WO@k}} &= \frac{1}{|Q|}\sum_{w \in Q}\text{WO@k}(w), \\
\overline{\text{PO@k}} &= \frac{1}{|Q|}\sum_{w \in Q}\text{PO@k}(w).
\end{aligned}
\end{equation}

Both metrics range from $0$ to $1$, the higher values the better cross-modal consistency. 
WO@$k$ evaluates exact lexical agreement, while PO@$k$ provides a softer evaluation based on phonetic similarity, capturing alignment even when retrievals differ lexically but remain close in phonetic space.

\subsection{Implementation}
We implement  a heterogeneous GCN with a single convolution block of type SAGEconv for each sub-graph ($G_a$, $G_l$ and $G_{al}$). As mentioned in section \ref{subsubsec:model} we use \textit{Mean aggregator}~\cite{GraphSAGE}. The GCN delivers a vector with the same size than the text encoder ($dim_{text}$). We use a MLP to get a BCE loss relative to a link prediction as described in Figure~\ref{fig:gcn}. This dense neural network has 1 input layer of size $dim_{text}$ and 1 hidden layer of size $2\times dim_{text}$. For training, we divided the set of heterogeneous edges into two parts: training  and test sets, with a ratio of 80:20. We trained the model for $1000$ epochs, and we empirically set $\alpha=0.1$ and $\beta=0.1$.

The purpose of the test set is to evaluate the actual capacity of the model in terms of link prediction.
For the link prediction, the MLP is trained jointly with the the HGNN.
As a reminder, we choose to move the acoustic vectors towards the linguistic space, which means that the linguistic space is kept fixed. 
Metrics are provided for three different steps: \textit{Init} for initial representations (\textit{e.g.}, XLSR or Wav2Vec), \textit{Learned} for the learned representations after graph-based training, and \textit{LP} (Link Prediction) for representations further refined through a link prediction model that is used to compute similarities between linguistic and acoustic vectors.
In the \textit{Learned} step the evaluated representations are the one obtained at inference time by the GNN only, while in the \textit{LP} step the evaluated representations are obtained at inference by the whole link prediction pipeline (GNN + MLP).



In the end, our experimental protocol involves two different datasets (TIMIT and YD), two acoustic representations (Wav2Vec2 and XLSR), three linguistic representations (LabSE, articulatory and phonetic). 
Notice that Wav2Vec is trained only on English data, while XLSR is trained on multilingual data (except Yemba). Therefore, we did not use Wav2Vec for Yemba.
Two different conditions are experimented: the transductive condition for which the evaluated words are the ones used to train the HGNN, the inductive condition for which the evaluated words have not been seen by the graph during training.

\begin{table*}[!h]
\centering
\small
\caption{
Comparison between our approach (Learned HGNN variants) and SAMU-XLSR 
in transductive and inductive settings on TIMIT. 
Word and phonetic overlaps are reported as $\overline{\text{WO@5}}$ and $\overline{\text{PO@5}}$ in percentages }
\label{tab:samu_timit}
\begin{tabular}{lcccc}
\toprule
\textbf{Setting and pipeline} & $\overline{\text{WO@5}}$ (\%) & $\overline{\text{PO@5}}$ (\%) & \textbf{Inter-Sim} & \textbf{Intra-Sim} \\
\midrule
\multicolumn{5}{l}{\textbf{Transductive}} \\
 SAMU-XLSR & 17.0 & 29.4 & 0.77 & 0.87 \\
 XLSR+HGNN+LaBSE & 30.6 & 42.2 & 0.89 & 0.94 \\
 XLSR+HGNN+Articulatory & 23.8 & 44.8 & 0.29 & 0.98 \\
 XLSR+HGNN+Bag of phoneme & 9.4 & 27.8 & 0.23 & 0.98 \\
\midrule
\multicolumn{5}{l}{\textbf{Inductive Semantic}} \\
 SAMU-XLSR & 31.8 & 41.6 & --- & --- \\
 XLSR+HGNN+LaBSE & 26.2 & 29.2 & --- & --- \\
 XLSR+HGNN+Articulatory & 16.8 & 19.4 & --- & --- \\
 XLSR+HGNN+Bag of phoneme & 6.6 & 8.6 & --- & --- \\
\midrule
\multicolumn{5}{l}{\textbf{Inductive Phonetic}} \\
 SAMU-XLSR & 24.6 & 34.8 & --- & --- \\
 XLSR+HGNN+LaBSE & 12.0 & 15.4 & --- & --- \\
 XLSR+HGNN+Articulatory & 23.2 & 25.2 & --- & --- \\
 XLSR+HGNN+Bag of phoneme & 7.6 & 9.8 & --- & --- \\
\bottomrule
\end{tabular}
\end{table*}

\section{Results and Discussions}
\label{RD}
\subsection{Comparison between our approach and SAMU-XLSR}

Table \ref{tab:samu_HGNN} compares the computational requirements and training configurations of the large-scale SAMU-XLSR model with our proposed lightweight approach based on XLSR combined with a heterogeneous graph neural network (HGNN). While SAMU-XLSR requires 300 million trainable parameters, extensive GPU resources (32 V100–32GB GPUs), and a massive 6.8 K hours dataset trained over 400K iterations, our method (XLSR + HGNN + LaBSE) operates with only 9 million parameters, runs on 32 CPUs, and can be trained in a matter of minutes to a few hours depending on the dataset (23 minutes for TIMIT and 2 hours for Yemba) for 1K training iterations. This illustrates the significant gap in computational cost and training time achieved by our approach.


\begin{table}[!h]
\centering
\tiny
\caption{Comparison of Computational Resources and Training Requirements between SAMU-XLSR and Our Approach}
\label{tab:samu_HGNN}
\begin{tabular}{lcc}
\toprule
\textbf{Metric} & \textbf{SAMU-XLSR} & \textbf{Our Approach} \\
\midrule
Train parameters & $300M$ & $9M$ \\
Hardware & 32 V100-32GB GPUs & 32 CPUs \\
Train Data  & 6.8 Kh & 23 min (TIMIT) \\
  &  & 2h (YD) \\
Train iterations & 400K & 1K  \\
\bottomrule
\end{tabular}
\end{table}

Tables \ref{tab:samu_timit} and \ref{tab:samu_yemba} present a comparison between SAMU-XLSR and our proposed approach across metrics of enrichment (inter- and intra-similarity) which evaluate the cohesion within the same word class and the separation across different word classes, respectively and cross-modal alignment: the Average word overlap ($\overline{\text{WO@5}}$) and phonetic overlap ($\overline{\text{PO@5}}$) on different experimental conditions. Our approach consistently outperforms SAMU-XLSR on the Yemba dataset across all evaluated metrics, demonstrating its effectiveness. Although results are comparable under transductive settings, the performance gap becomes substantial in inductive settings, where SAMU-XLSR fails to identify any matches at top@k=5. This disparity could be attributed to the absence of Yemba in both the language sets covered by XLSR and LaBSE.

\begin{table*}[!h]
\centering
\small
\caption{
Comparison between our approach (Learned HGNN variants) and SAMU-XLSR 
in transductive and inductive settings on the Yemba dataset. 
Word and phonetic overlaps are reported as $\overline{\text{WO@5}}$ and $\overline{\text{PO@5}}$ in percentages}
\label{tab:samu_yemba}
\begin{tabular}{lcccc}
\toprule
\textbf{Setting and pipeline} & $\overline{\text{WO@5}}$ (\%) & $\overline{\text{PO@5}}$ (\%) & \textbf{Inter-Sim} & \textbf{Intra-Sim} \\
\midrule
\multicolumn{5}{l}{\textbf{Transductive}} \\
SAMU-XLSR & 13.6 & 22.0 & 0.67 & 0.82 \\
XLSR+HGNN+LaBSE & 20.2 & 33.2 & 0.26 & 0.98 \\
XLSR+HGNN+Articulatory & 12.6 & 26 & 0.16 & 0.98 \\
XLSR+HGNN+Bag of phonemes & 22.2 & 38.6 & 0.31 & 0.98 \\
\midrule
\multicolumn{5}{l}{\textbf{Inductive}} \\
SAMU-XLSR & 0.0 & 0.0 & --- & --- \\
XLSR+HGNN+LaBSE & 20.0 & 20.0 & --- & --- \\
XLSR+HGNN+Articulatory & 11.8 & 11.8 & --- & --- \\
XLSR+HGNN+Bag of phonemes & 18.8 & 24 & --- & --- \\
\bottomrule
\end{tabular}
\end{table*}

\begin{figure*}[ht]
    \centering
    
    \begin{subfigure}[b]{0.48\textwidth}
        \begin{tikzpicture}
            \begin{axis}[
                title={},
                xlabel={$k$},
                ylabel={Average word Overlap (\%)},
                xmin=5, xmax=25,
                ymin=0, ymax=70,
                xtick={5,10,15,20,25},
                ytick={0,10,20,30,40,50,60,70},
                legend pos=north west,
                grid=major,
                width=\textwidth,
                height=0.8\textwidth]
                
                \addplot[blue, mark=square] coordinates {
                    (5,31.8)
                    (10,35.8)
                    (15,37.13)
                    (20,40.00)
                    (25,43.64)
                };
                \addlegendentry{SAMU-XLSR}
                
                \addplot[red, mark=triangle] coordinates {
                    (5,25.00)
                    (10,29.10)
                    (15,32.40)
                    (20,35.40)
                    (25,39.04)
                };
                \addlegendentry{Our Approach}
            \end{axis}
        \end{tikzpicture}
        \caption{TIMIT-Semantic Dataset}
    \end{subfigure}
    \hfill
    \begin{subfigure}[b]{0.48\textwidth}
        \begin{tikzpicture}
            \begin{axis}[
                title={},
                xlabel={$k$},
                ylabel={Average word Overlap (\%)},
                xmin=5, xmax=25,
                ymin=0, ymax=70,
                xtick={5,10,15,20,25},
                ytick={0,10,20,30,40,50,60,70},
                legend pos=north west,
                grid=major,
                width=\textwidth,
                height=0.8\textwidth]
                
                \addplot[blue, mark=square] coordinates {
                    (5,0.00)
                    (10,11.20)
                    (15,20.80)
                    (20,40.00)
                    (25,40.48)
                };
                \addlegendentry{SAMU-XLSR}
                
                \addplot[red, mark=triangle] coordinates {
                    (5,20.00)
                    (10,29.40)
                    (15,32.53)
                    (20,39.70)
                    (25,48.00)
                };
                \addlegendentry{Our Approach}
            \end{axis}
        \end{tikzpicture}
        \caption{Yemba Dataset}
    \end{subfigure}
       
    \caption{Performance comparison between SAMU-XLSR and our approach XLSR + HGNN + LaBSE, across TIMIT and Yemba inductive dataset on word overlap metric  at varying top-k values}
    \label{fig:full_comparison}
\end{figure*}
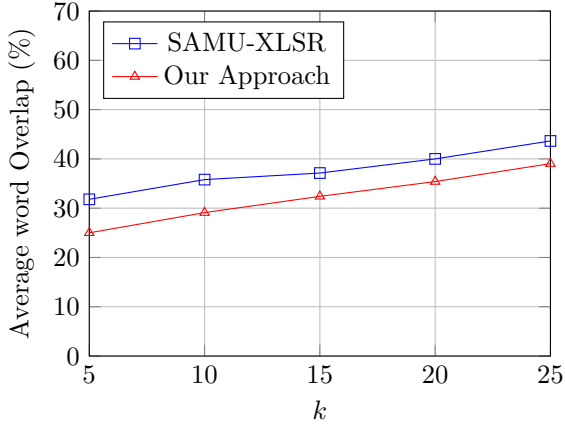
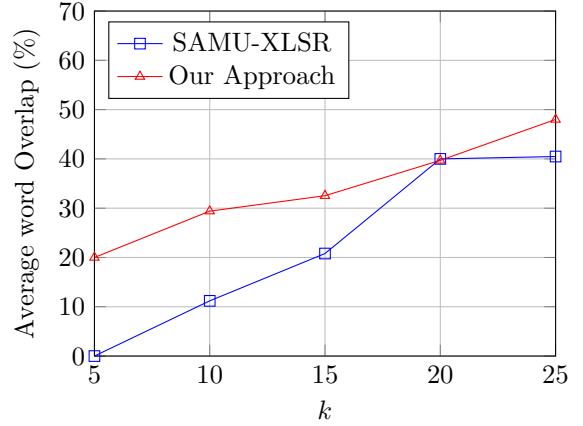

This observation is further corroborated by results on the TIMIT dataset, where SAMU-XLSR exhibits superior performance and outperforms our approach in inductive settings. However, intra- and inter-similarity metrics remain favourable to our approach, particularly when employing phonetic-based representations. Figure \ref{fig:full_comparison} illustrates the average word overlap metrics comparing SAMU-XLSR with our approach (XLSR + HGNN + LaBSE) across different top@k values. The results indicate that when considering semantic cross-modal word alignment alone, our approach achieves comparable performance to SAMU-XLSR on TIMIT and overcome it on Yemba datasets under inductive settings, even by expanding the neighbourhood. Although SAMU-XLSR shows slightly better overall performance in English (its training language) on TIMIT, Table \ref{tab:samu_HGNN} clearly illustrates the advantages of the frugality and interpretability of the approach we proposed.

Futhermore, while SAMU-XLSR primarily focuses on semantic alignment, leveraging articulatory representations as a teacher enables the model to emphasize phonetic alignment. This approach enhances the interpretability of the learned acoustic representations and improves their intrinsic quality. Moreover, unlike neural representations such as LaBSE, articulatory features are handcrafted, easily extractable across languages, and thus particularly well-suited for low-resource language scenarios.


\subsection{Impact of the initial representations}
\subsubsection{Enrichment results}

Table \ref{tab:inter_intra_similarity_only} summarizes intra-similarity and inter-similarity metrics, before(\textit{init}) and after the alignment(Learned/+LP) for each text and acoustic representation and for each dataset. Note that \textit{Learned} and \textit{LP} share the same acoustic vectors, and thus have identical intra/inter similarity values. These similarity metrics are measured on acoustic vectors only and are relevant for both transductive and inductive settings

\begin{table*}[!h]
\centering
\small
\caption{Intra- ($\uparrow$) and inter-similarity ($\downarrow$) under transductive and inductive settings on TIMIT dataset (Wav2Vec, XLSR) and Yemba (XLSR)}
\label{tab:inter_intra_similarity_only}
\begin{tabular}{lcccccc}
\toprule
\textbf{Setting} & \multicolumn{2}{c}{\textbf{LaBSE}} & \multicolumn{2}{c}{\textbf{Articulatory}} & \multicolumn{2}{c}{\textbf{Bag of phonemes}}\\
& Init & Learned/+LP & Init & Learned/+LP & Init & Learned/+LP\\
\midrule
\multicolumn{7}{l}{\textbf{TIMIT Wav2vec}} \\
Intra-similarity & 0.78 & 0.95 & 0.78 & \textbf{0.96} & 0.78 & 0.93 \\
Inter-similarity & 0.70 & \textbf{0.18} & 0.70 & 0.25 & 0.70 & 0.23 \\
\midrule
\multicolumn{7}{l}{\textbf{TIMIT XLSR}} \\
Intra-similarity & \textbf{0.98} & 0.97 & \textbf{0.98} & \textbf{0.98} & \textbf{0.98} & 0.96 \\
Inter-similarity & 0.98 & 0.27 & 0.98 & 0.29 & 0.97 & \textbf{0.24} \\
\midrule
\multicolumn{7}{l}{\textbf{Yemba XLSR}} \\
Intra-similarity & 0.98 & 0.98 & 0.98 & 0.98 & 0.98 & 0.98 \\
Inter-similarity & 0.98 & 0.26 & 0.98 & \textbf{0.16} & 0.98 & 0.31 \\
\bottomrule
\end{tabular}
\end{table*}

The results reported in Table \ref{tab:inter_intra_similarity_only} consistently demonstrate that, regardless of the dataset or the type of linguistic and embedding employed, the intra-similarity of the \textit{Learned} and \textit{LP} representations systematically equals or exceeds that of the initial representations. Moreover, the inter-similarity scores show a significant improvement in the learned/LP configurations compared to their initial counterparts. Although the initial representations may already capture intra-word similarity to some extent, our approach substantially enhances the separability of representations across different words. On the Yemba dataset, the best inter-similarity score of $0.16$ is achieved using the articulatory representation, while on TIMIT, the best score is obtained with the bag-of-phonemes representation when using XLSR as the acoustic representation and LaBSE when using wav2vec. This shows that phonetic representations provide a comparable level of enrichment to LaBSE on TIMIT, while outperforming it on the Yemba dataset. These findings highlight the potential of phonetic enrichment over semantic embeddings such as LaBSE for low-resource languages and demonstrate the ability of our HGNN model to enhance the intrinsic quality of acoustic representations regardless of the text representation used.

\subsubsection{Word alignment results}

Tables~\ref{tab:wav2vec_overlap_timit}, \ref{tab:xlsr_overlap_timit}, and \ref{tab:xlsr_overlap_yemba} report Average word overlap ($\overline{\text{WO@5}}$) and phonetic overlap ($\overline{\text{PO@5}}$) on different experimental conditions.
All results are based on the top 5 nearest neighbours.

\begin{table*}[!h]
\centering
\small
\caption{
Average word overlap ($\overline{\text{WO@5}}$, $\uparrow$) and phonetic overlap ($\overline{\text{PO@5}}$, $\uparrow$) 
under transductive and inductive settings on the TIMIT dataset (using Wav2Vec embeddings). 
Results are reported with $k=5$. 
Values are percentages, where 100\% corresponds to perfect retrieval.
}
\label{tab:wav2vec_overlap_timit}
\begin{tabular}{lccccccccc}
\toprule
\textbf{Setting} & \multicolumn{3}{c}{\textbf{LaBSE}} & \multicolumn{3}{c}{\textbf{Articulatory}} & \multicolumn{3}{c}{\textbf{Bag of phonemes}} \\
& Init & Learned & +LP & Init & Learned & +LP & Init & Learned & +LP \\
\midrule
\multicolumn{10}{l}{\textbf{Transductive}} \\
$\overline{\text{WO@5}}$ (\%) & 6.4 & 29.2 & 27.8 
& 3.6 & 27.4 & \textbf{31.2} 
& 5.0 & 13.0 & 8.6 \\
$\overline{\text{PO@5}}$ (\%) & 17.6 & 39.8 & 40.0 
& 10.0 & 47.4 & \textbf{54.2} 
& 11.6 & 37.4 & 21.2 \\
\midrule
\multicolumn{10}{l}{\textbf{Inductive (Semantic)}} \\
$\overline{\text{WO@5}}$ (\%) & 5.0 & \textbf{26.6} & 19.4 
& 9.0 & 19.0 & 17.6 
& 5.6 & 7.2 & 8.8 \\
$\overline{\text{PO@5}}$ (\%) & 5.6 & \textbf{29.6} & 22.8 
& 6.0 & 20.0 & 21.0 
& 13.6 & 10.0 & 12.2 \\
\midrule
\multicolumn{10}{l}{\textbf{Inductive (Phonetic)}} \\
$\overline{\text{WO@5}}$ (\%) & 3.6 & 14.0 & 9.0 
& 4.8 & \textbf{28.8} & 23.2 
& 6.2 & 10.8 & 6.6 \\
$\overline{\text{PO@5}}$ (\%) & 6.2 & 18.2 & 13.6 
& 4.6 & \textbf{30.2} & 26.2 
& 6.6 & 14.2 & 9.6 \\
\bottomrule
\end{tabular}
\end{table*}

\begin{table*}[!h]
\centering
\small
\caption{
Average word overlap ($\overline{\text{WO@5}}$, $\uparrow$) and phonetic overlap ($\overline{\text{PO@5}}$, $\uparrow$) 
under transductive and inductive settings on the TIMIT dataset (using XLSR embeddings). 
Results are reported with $k=5$. 
Values are percentages, where 100\% corresponds to perfect retrieval.
}
\label{tab:xlsr_overlap_timit}
\begin{tabular}{lccccccccc}
\toprule
\textbf{Setting} & \multicolumn{3}{c}{\textbf{LaBSE}} & \multicolumn{3}{c}{\textbf{Articulatory}} & \multicolumn{3}{c}{\textbf{Bag of phonemes}} \\
& Init & Learned & +LP & Init & Learned & +LP & Init & Learned & +LP \\
\midrule
\multicolumn{10}{l}{\textbf{Transductive}} \\
$\overline{\text{WO@5}}$ (\%) & 2.8 & \textbf{30.0} & 29.2 
& 3.6 & 26.6 & 26.8 
& 5.8 & 11.0 & 8.2 \\
$\overline{\text{PO@5}}$ (\%) & 7.0 & 41.2 & 40.8 
& 10.2 & \textbf{47.2} & 43.4 
& 15.0 & 33.8 & 22.8 \\
\midrule
\multicolumn{10}{l}{\textbf{Inductive (Semantic)}} \\
$\overline{\text{WO@5}}$ (\%) & 4.6 & \textbf{25.0} & 20.2 
& 6.2 & 15.6 & 14.6 
& 6.8 & 11.4 & 10.4 \\
$\overline{\text{PO@5}}$ (\%) & 4.8 & \textbf{27.2} & 22.2 
& 6.4 & 18.2 & 16.4 
& 7.6 & 15.0 & 12.2 \\
\midrule
\multicolumn{10}{l}{\textbf{Inductive (Phonetic)}} \\
$\overline{\text{WO@5}}$ (\%) & 3.2 & 12.4 & 9.2 
& 3.2 & \textbf{24.2} & 20.0 
& 5.2 & 12.2 & 7.0 \\
$\overline{\text{PO@5}}$ (\%) & 3.2 & 16.4 & 12.8 
& 4.0 & \textbf{26.0} & 21.6 
& 7.0 & 16.8 & 9.0 \\
\bottomrule
\end{tabular}
\end{table*}

\begin{table*}[!h]
\centering
\small
\caption{
Average word overlap ($\overline{\text{WO@5}}$, $\uparrow$) and phonetic overlap ($\overline{\text{PO@5}}$, $\uparrow$) 
under transductive and inductive settings on the Yemba dataset (using XLSR embeddings). 
}
\label{tab:xlsr_overlap_yemba}
\begin{tabular}{lccccccccc}
\toprule
\textbf{Setting} & \multicolumn{3}{c}{\textbf{LaBSE}} & \multicolumn{3}{c}{\textbf{Articulatory}} & \multicolumn{3}{c}{\textbf{Bag of phonemes}} \\
& Init & Learned & +LP & Init & Learned & +LP & Init & Learned & +LP \\
\midrule
\multicolumn{10}{l}{\textbf{Transductive}} \\
$\overline{\text{WO@5}}$ (\%) & 5.6 & 20.2 & 16.2 
& 7.6 & 12.6 & 13.2 
& 9 & 22.2 & \textbf{24.0} \\
$\overline{\text{PO@5}}$ (\%) & 12.2 & 33.2 & 21.0 
& 17.0 & 26.0 & 21.6 
& 17.6 & 38.6 & \textbf{34.6} \\
\midrule
\multicolumn{10}{l}{\textbf{Inductive}} \\
$\overline{\text{WO@5}}$ (\%) & 2.4 & \textbf{20.0} & 18.8 
& 3.2 & 11.8 & 5.2 
& 4.8 & 18.8 & 16.4 \\
$\overline{\text{PO@5}}$ (\%) & 2.4 & 20.0 & 18.8 
& 3.2 & 11.8 & 5.2 
& 5.4 & 24.0 & \textbf{22.0} \\
\bottomrule
\end{tabular}
\end{table*}

The previous  observations on enrichment results are consistent with the cross-modal alignment metrics observable in Tables~\ref{tab:wav2vec_overlap_timit}, \ref{tab:xlsr_overlap_timit}, and \ref{tab:xlsr_overlap_yemba}. Indeed, on the TIMIT dataset (Tables \ref{tab:wav2vec_overlap_timit} and \ref{tab:xlsr_overlap_timit}), among the different linguistic embeddings, the articulatory-based representation achieves the best performance in transductive settings (\textit{Learned} and \textit{LP} configurations), followed by the LaBSE representation. In inductive settings, we observe that articulatory representations perform best on phonetic datasets, while LaBSE achieves superior performance on semantic tasks. This pattern remains consistent across both Wav2Vec and XLSR implementations.
These results align with theoretical expectations, as LaBSE is fundamentally a semantic representation while articulatory features are inherently phonetic. This confirms the relevance of articulatory representations for phonological alignment tasks compared to semantic representations such as LaBSE. The metrics are generally better with Wav2vec than with XLSR. This observation can be explained by the fact that the Wav2Vec integration is specialised in English, so the initial acoustic representation is better than that provided by XLSR.

For the Yemba dataset (Table \ref{tab:xlsr_overlap_yemba}), optimal performance is achieved by bag-of-phonemes in the transductive setting; in the inductive setting, it shares first place with LaBSE, which is the best in terms of average word overlap, which corresponds to the observations made on TIMIT, namely that phonetic representation gives the best results for phonetic alignment. The absolute performance values on the Yemba dataset show a considerable degradation compared to those obtained on TIMIT, which could be attributed to the lower quality of initial vector representations produced by XLSR for this language. XLSR is mainly trained with English data, which explains the comparable results between Tables \ref{tab:wav2vec_overlap_timit} and \ref{tab:xlsr_overlap_timit}. However, Yemba is a low-resource language not included in XLSR's training corpus. Consequently, it might be possible that the generated embeddings for Yemba are inherently approximate.
This observation is further reinforced by the very low score obtained by the \textit{Init} configuration in inductive parameters, regardless of the text representation method. Nevertheless, these results demonstrate the robustness of the proposed method, which maintains effectiveness even when initialized with low-quality representations.

Although the overall performance shows some degradation, the inter- and intra-similarity metrics continue to favor the learned representations, with particularly notable improvements in inter-similarity measures.
These results suggest that although cross-modal alignment effectiveness is somewhat reduced, the HGNN and its message-passing algorithm still facilitate substantial representational enrichment, with particularly pronounced benefits observed for phonetic text representation method. Since our method does not incorporate fine-tuning of the encoder, those results also indicate that the graph-based learning framework maintains its capacity to enhance representation quality even without encoder optimization, demonstrating the robustness of the proposed approach.


\section{Conclusion}
  \label{sec:con}

This paper presents an efficient and novel approach to cross-modal alignment for enriching speech representations in low-resource language settings, particularly under computational resource constraints. Our core insight is that cross-modal alignment can be formulated as a link prediction task within a heterogeneous graph neural network framework, while acoustic representations are simultaneously enriched through the integration of linguistic information. Our primary contribution thus lies in using heterogeneous graphs to merge both modalities and facilitate information exchange between them via message-passing mechanisms.

To assess the effectiveness of our approach, experiments were conducted on two datasets: TIMIT (English) and Yemba (a Cameroonian language). For each dataset, we aligned speech and text representations at the word level using different text and speech encoders. Specifically, we introduced two phonetic representations to enhance phonetic alignment: an articulatory-based representation and a bag-of-phonemes representation.

Performance was evaluated using four metrics: average word overlap and average phonetic overlap to assess cross-modal alignment quality, and inter- and intra-similarity measures to evaluate representation enrichment. We compared the embeddings obtained with our approach to those from SAMU-XLSR, a state-of-the-art method. In terms of inter- and intra-similarity, our approach yields greater improvements in the intrinsic monolingual properties of the embeddings than SAMU-XLSR. Alignment performance was comparable on English and superior on Yemba. While matching the performance of SAMU-XLSR, our approach requires substantially less training data and computational resources — directly addressing the primary objective of this work.

We further compared the initial embeddings with those obtained after alignment to quantify the extent to which our approach enriches speech representations. Our method consistently improved both enrichment and alignment metrics for the acoustic representations derived from Wav2Vec and XLSR, across both datasets and regardless of the text representation used. As expected, articulatory-based representations proved more effective for phonetic retrieval, while LaBSE performed better for semantic retrieval — highlighting the distinct type of enrichment each text representation contributes.

These promising results open up several directions for future work. First, experiments could be extended to the sentence level, which introduces the added complexity of handling longer sequential dependencies. Second, it would be valuable to assess the impact of this approach within end-to-end pipelines for downstream speech tasks, such as speech synthesis and speech recognition, to evaluate its practical utility in real-world applications.


\section*{Declarations}


\subsection*{Availability of data and materials} 
The datasets generated and/or analysed during the current study are available in the repositories:
\begin{itemize}
    \item TIMIT : \url{https://www.kaggle.com/mfekadu/darpa-timit-acousticphonetic-continuous-speech}
    \item Yemba : \url{https://doi.org/10.17632/74p9d5frg3.1}
\end{itemize}

\subsection*{Competing interests} 
The authors declare that they have no competing interests.


\subsection*{Funding} 
This work has been funded by the European Union’s Horizon 2020 research and
innovation program under the Marie Skłodowska-Curie grant agreement No 101007666.

\subsection*{Authors' contributions}
\begin{itemize}
    \item Yannick Yomie Nzeuhang: conceptualization, data (pre-)processing, experiments, writing original draft.
    \item Marie Tahon: methodology, conceptualization, supervision, results interpretation, writing-reviewing.
    \item Paulin Melatagia Yonta: methodology,results interpretation, writing-reviewing.
\end{itemize}

\subsection*{Acknowledgments}
The authors also thanks the LIUM for its computing ressources.

\subsection*{Code availability}
The code is available free of charge and open access on GitHub, and the link is provided in the paper.






\begin{appendices}

\section{}
\label{sec:ann1}

The Yemba is an African language spoken primarily in the Menoua region of western Cameroon, part of the Grassfields group. The name “Yémba,” meaning “I say that”; Yemba belongs to the Bantoid language family, specifically the Southern Bantoid subgroup within the Niger-Congo languages, which are classified under the hypothetical Voltaic-Congo family \cite{greenberg1963languages}.

The Yemba alphabet consists of 33 letters and adheres to the General Alphabet of Cameroonian Languages (GACL), published in 1978 and officially adopted in 1984 by the National Association of Cameroonian Language Committees (NAALC) \cite{bird1997petit}. Linguistically, Yemba is notable for it relatively distinct tonal system. As a tonal language, Yemba employs high, medium, and low tones, with pitch variations that critically influence word meanings even when consonants and vowels remain the same \cite{bird1997petit}. This tonal structure, common among Bantu languages, adds a layer of complexity to its spoken form, making pitch an essential component of Yemba pronunciation.   The table \ref{tab:yemba} presents some word of the dataset.

\begin{table}[!h]

\centering
\begin{tabular}{|c|c|}
\hline
\textbf{Term} & \textbf{English Translation} \\

\textipa{mbeN} & The rain \\

\textipa{mb\=iN} & The forest\\

\textipa{lekwEt} & A mountain \\
    
 \textipa{mbuO} & The chalk \\
   
met\'ua & A car\\

\textipa{kamElE} & The vaccination\\

\textipa{mbalON} & A ball\\

\textipa{akia} & A graduated ruler\\

\textipa{nu} & The sun \\

\textipa{ashuNO} & The driving \\

\textipa{Ng\=ap} & The hen \\

\textipa{saN} & The moon \\

\textipa{nzenzhE} & A fly \\
\hline

\end{tabular}
\caption{Terms and their English Translations}
\label{tab:yemba}
\end{table}

\section{}

This table\ref{tab:correspondances} presents the 109 words extract from TIMIT alongside their semantic equivalents (words with similar meanings) and phonetic equivalents (words that sound alike but have different meanings).

\begin{table}[h]
\centering
\begin{tabular}{|c|c|c|}
\hline
\textbf{Original} & \textbf{Semantic} & \textbf{Phonetic} \\
\hline
much & lots & match \\
near & close & kneer \\
often & frequently & orphan \\
please & kindly & fleece \\
problem & issue & propel \\
said & stated & sad \\
saw & seen & sore \\
take & grab & ache \\
them & those people & thumb \\
\hline
\end{tabular}
\caption{Correspondences between original words, their semantic and phonetic equivalents}
\label{tab:correspondances}
\end{table}




\end{appendices}

\newpage
\bibliography{sn-bibliography}

\end{document}